\documentclass[10pt,twocolumn,letterpaper]{article}

\usepackage{cvpr}
\usepackage{amsmath,amssymb,booktabs,tabularx,xspace}
\usepackage{fix-cm}
\usepackage{microtype}
\usepackage{pifont}

\makeatletter
\def\NAT@cmprs{0}
\makeatother

\AddToHook{env/equation/begin}{%
  \addtolength{\abovedisplayskip}{-2.5pt}%
  \addtolength{\belowdisplayskip}{-2.5pt}%
  \addtolength{\belowdisplayshortskip}{-2.5pt}%
}

\makeatletter
\renewcommand\paragraph{\@startsection{paragraph}{4}{\z@}
  {5pt}{-1em}{\normalfont\normalsize\bfseries}}
\makeatother

\newcommand{\method}{Paste2Place\xspace}

\definecolor{cvprblue}{rgb}{0.21,0.49,0.74}
\usepackage[pagebackref,breaklinks,colorlinks,allcolors=cvprblue]{hyperref}

\def\confName{CVPR}
\def\confYear{2027}

\title{A Bridge from Paste to Place:\\
Physics-Guided One-Step Image Composition via Object--Scene Interaction}
\newlength{\authornamewidth}
\author{%
\makebox[\authornamewidth][c]{Hongzhen Li\rlap{$^{\dagger}$}}\hspace{1.5em}%
\makebox[\authornamewidth][c]{Youwei Pan}\hspace{1.5em}%
\makebox[\authornamewidth][c]{Yingfang Zhu}\\
\makebox[\authornamewidth][c]{Sen Zhu}\hspace{1.5em}%
\makebox[\authornamewidth][c]{Fengjie Zhu}\hspace{1.5em}%
\makebox[\authornamewidth][c]{Leilei Cao\rlap{$^{*}$}}\\[8pt]
Transsion%
}
\hypersetup{pdfauthor={Hongzhen Li, Youwei Pan, Yingfang Zhu, Sen Zhu, Fengjie Zhu, Leilei Cao},
  pdftitle={A Bridge from Paste to Place: Physics-Guided One-Step Image Composition via Object-Scene Interaction}}

\makeatletter
\patchcmd{\@maketitle}{\null}{\null\kern-20pt}{}%
  {\PackageError{preprint}{Title top spacing patch failed}{}}
\patchcmd{\@maketitle}{\vskip .5em}{\vskip 1pt}{}%
  {\PackageError{preprint}{Author spacing patch failed}{}}
\patchcmd{\@maketitle}{\vspace*{12pt}}{\vspace*{0pt}}{}%
  {\PackageError{preprint}{Title ending spacing patch failed}{}}

\def\ps@authornotes{%
  \let\@oddhead\@empty\let\@evenhead\@empty
  \def\@oddfoot{\normalfont\footnotesize
    \parbox[b]{\textwidth}{%
      \rule{0.4\columnwidth}{0.4pt}\\[3pt]
      $^{\dagger}$Work done as an intern at Transsion.\\
      $^{*}$Corresponding author.}}%
  \let\@evenfoot\@oddfoot
}
\patchcmd{\abstract}{\thispagestyle{empty}}{\thispagestyle{authornotes}}{}%
  {\PackageError{preprint}{Author footer patch failed}{}}
\makeatother

\AddToHook{cmd/@maketitle/after}{\noindent
\begin{minipage}{\textwidth}
    \centering
    \includegraphics[width=\linewidth]{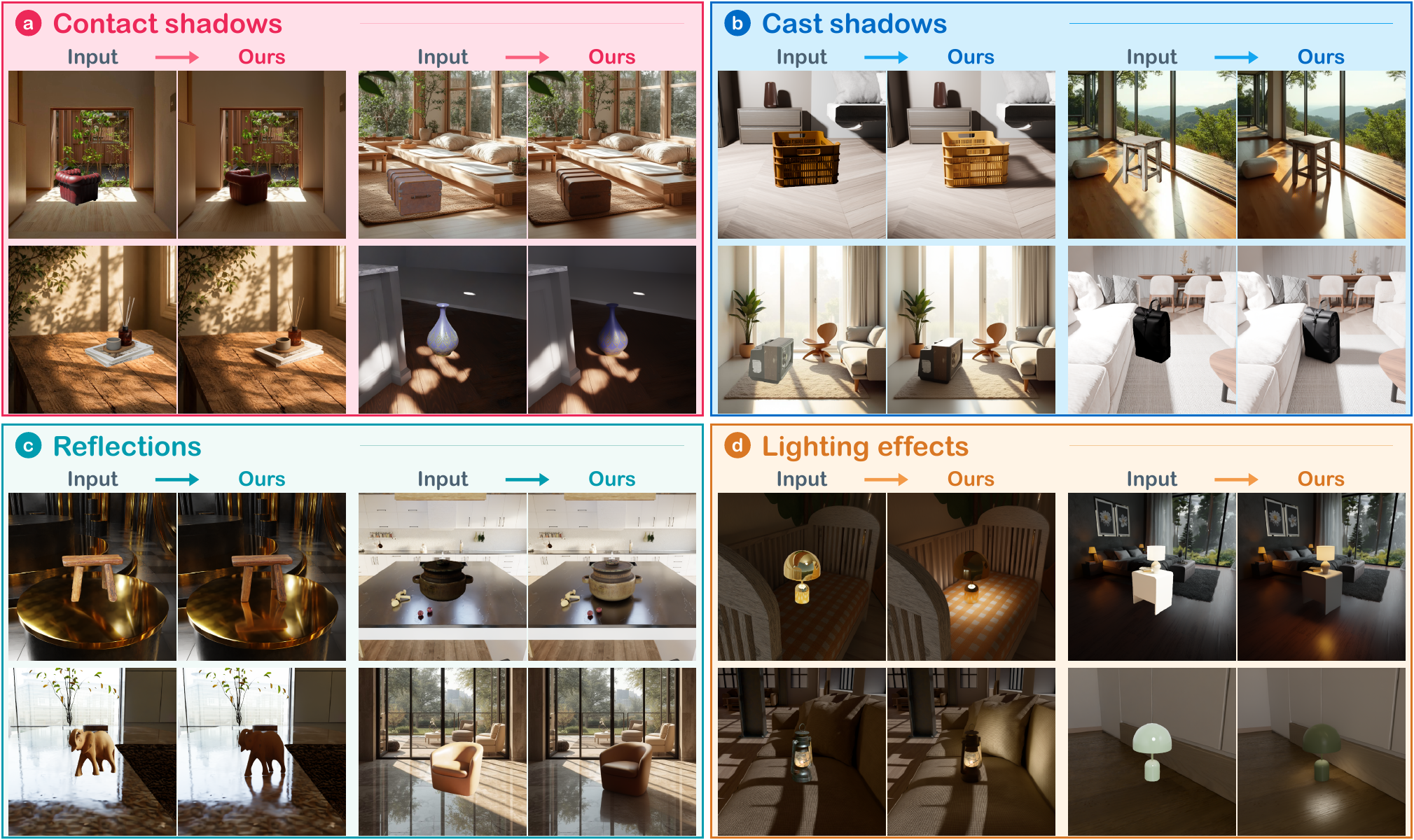}
    \captionsetup{hypcap=false}
    \captionof{figure}{\textbf{Realistic image composition in one step.}
    \method makes pasted objects appear as if they were physically placed in the scene, turning ``paste'' into ``place.''
    Examples are grouped by effect: (a) contact shadows, (b) cast shadows, (c) reflections, and (d) lighting effects.}
    \label{fig:teaser}
\end{minipage}\par
\vspace{18pt}
}

\patchcmd{\endabstract}{\vspace*}{\vspace}{}{\errmessage{Abstract patch failed}}
\AddToHook{env/abstract/end}{\looseness=-1\par}

\begin{document}
\maketitle
\begin{abstract}
Realistic image composition makes a pasted object appear naturally present in a scene while preserving the pasted layout and fine foreground and background details.
Reference-based insertion methods preserve overall object identity but can alter pose and fine details, often at substantial computational cost.
Foreground harmonization and shadow generation address individual aspects of this task, but neither alone ensures natural integration.
An efficient, unified solution remains an open challenge, and large-scale, high-quality datasets and benchmarks for this task are lacking.
To address these challenges, we introduce \method, an end-to-end, physics-guided flow-based model that learns a direct latent bridge from pasted inputs to realistic composites, guided by image-space corrections analytically derived from predicted physical components of object and scene appearance changes.
To support training and evaluation, we introduce P2P-Data with 86K manually reviewed synthetic and real-scene training pairs, including dense physical annotations from our Blender rendering and validation pipeline for the synthetic subset.
P2P-Bench combines paired compositions and manually pasted inputs from collected images into 1.5K examples for evaluating composition under user-chosen placements.
Extensive qualitative and quantitative experiments demonstrate that \method achieves substantial advantages in image composition quality and inference speed across diverse scenarios.
\end{abstract}

\section{Introduction}
\label{sec:intro}

Image composition is a key task in visual content creation, aiming to make a pasted object appear naturally present in a scene.
It has broad applications in photo editing, advertising, and visual effects.
Given an object cropped from one image and pasted onto another, the task is to turn ``paste'' into ``place'': making the object look physically present while preserving the user-specified layout and fine foreground and background details (Fig.~\ref{fig:teaser}).

\begin{figure}[t]
    \centering
    \includegraphics[width=\columnwidth]{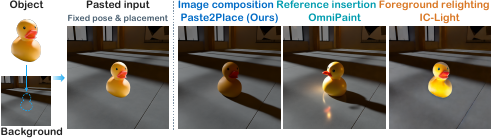}
    \caption{\textbf{Different objectives in object placement.}
    Visual comparison of three related but distinct tasks: image composition, reference-based insertion, and foreground relighting.}
    \label{fig:task-definition}
\end{figure}

Achieving this goal reliably and efficiently remains challenging for existing methods.
Reference-based insertion~\cite{song2026insert,chen2024anydoor,yu2025omnipaint} and general image editing~\cite{wu2025qwenimage,mao2025aceplus} can adapt an object to a new scene, but may also change its pose or regenerate fine details.
Foreground harmonization and relighting~\cite{jin2024neuralgaffer,ren2024relightful,zhang2025iclight,careaga2023intrinsic,yu2024cfdiffusion} and shadow generation~\cite{li2026vsdiffusion,wang2025metashadow,zhao2025gpsdiffusion,liu2024sgdiffusion} address selected aspects of integration, but do not fully capture realistic physical interactions between the foreground object and its surrounding scene.
In Fig.~\ref{fig:task-definition}, OmniPaint~\cite{yu2025omnipaint} changes the pasted pose, whereas IC-Light~\cite{zhang2025iclight} relights the object but leaves its cast shadow absent.

\suppressfloats[t]

These gaps lead to four main challenges.
\ding{182}~\textbf{Content preservation.} Generative changes can alter a pasted object's pose, position, or identity and erode fine details, such as intricate patterns and small text, in both the object and its background.
\ding{183}~\textbf{Realistic physical interaction.} A newly placed object must respond to the scene's illumination while also changing the light reaching surrounding surfaces.
Different materials and illumination conditions produce different effects ranging from contact and cast shadows to reflections and lighting effects (Fig.~\ref{fig:teaser}).
\ding{184}~\textbf{Efficiency.} Expensive iterative inference keeps users waiting as they repeatedly adjust and compare placements, disrupting interactive creation.
\ding{185}~\textbf{Data.} Large-scale, high-quality datasets and benchmarks remain scarce for this task, especially with physical annotations across diverse interactions.

We develop a physics-guided flow-based framework \emph{\method}, in which we introduce the \emph{Source-Anchored Latent Bridge} to address challenges \ding{182} and \ding{184}.
The pasted image already specifies what the user wants to retain: the object, its arrangement, and the scene detail.
Its target differs mainly in the appearance changes needed for integration, so the two endpoints share substantial content.
Rather than reconstructing this content through repeated denoising from Gaussian noise, we start from the pasted image's latent representation.
The input is thus the state being transformed, not merely a condition on generation.
We train this mapping end to end and predict the full change between aligned endpoints with one DiT evaluation at inference.
Source anchoring helps preserve layout and detail; the single-step update avoids iterative sampling and gives users faster feedback.

To address challenge \ding{183}, we design \emph{Physics-Based Appearance Change Modeling} in \method.
A direct source-to-target mapping alone does not distinguish a surface's material appearance from the influence of illumination.
Our central insight is that making an object appear placed requires modeling both how the scene illuminates it and how its presence alters light transport around it.
An albedo--shading--residual representation~\cite{careaga2023intrinsic,liang2026pilight,careaga2024colorful,careaga2025physicalrelighting} separates the surface's intrinsic diffuse color (albedo) from illumination-dependent shading and residual effects, including reflections and emission.
We predict object albedo and these physical changes, then analytically combine them into object and environment corrections.
These corrections guide the latent update within the same forward pass, making physical prediction part of image generation rather than a separate decomposition task.

Addressing challenge \ding{185} requires content-aligned training pairs and physical supervision for the appearance changes.
We design a Blender rendering and validation pipeline to create synthetic pairs with dense physical annotations, and construct real-scene composition pairs from photographs in the OBER object-removal dataset~\cite{zhao2026objectclear}.
After rigorous manual review, we assemble the \emph{Paste2Place Dataset} (\emph{P2P-Data}) with 86{,}273 training pairs.
We also construct the \emph{Paste2Place Benchmark} (\emph{P2P-Bench}), comprising 1{,}500 rendered, real-scene, and manually pasted examples from collected images for evaluating paired reconstruction quality and composition under user-chosen arrangements.
Both resources will be released to support further research.

Quantitative comparisons, visual examples, and human ratings demonstrate improved composition quality and detail preservation across diverse interactions.
By completing composition in a single forward pass, Paste2Place achieves substantially faster inference than iterative generation.

Our contributions can be summarized as follows:
\begin{itemize}[before=\interlinepenalty10000]
    \item We introduce \emph{\method}, a physics-guided flow-based framework whose \emph{Source-Anchored Latent Bridge} transforms pasted inputs into realistic composites directly.
    \item We propose \emph{Physics-Based Appearance Change Modeling}, which predicts physical components of object and environment appearance changes and analytically recomposes them into corrections that guide the latent update.
    \item We design a Blender rendering and validation pipeline and introduce \emph{P2P-Data}, comprising 86{,}273 training pairs with dense physical annotations for the synthetic subset.
    \item We establish \emph{P2P-Bench}, a 1{,}500-example benchmark spanning rendered, real-scene, and manually pasted composites for evaluating composition quality and efficiency.
\end{itemize}

\section{Related Work}
\label{sec:related}

\subsection{Image Relighting and Harmonization}
Image harmonization adjusts foreground appearance to match its background while retaining the composite structure~\cite{tsai2017harmonization}.
Learning-based methods align foreground and background feature distributions~\cite{cong2020dovenet,ling2021rain} or aggregate long-range background context~\cite{shen2023gknet,zhou2026wavssnet}.
High-resolution approaches learn color transformations and filters~\cite{ke2022harmonizer,xue2022dccf,guerreiro2023pctnet,meng2024aict} or combine pixel- and color-level predictions~\cite{cong2022cdtnet} to retain fine detail.
Generative approaches use diffusion priors to capture broader appearance variations through supervised adaptation or zero-shot optimization~\cite{zhou2024diffharmony,zhou2024diffharmonypp,chen2025zeroshotharmonization}.

Relighting addresses illumination more explicitly.
Intrinsic approaches separate reflectance from shading to adapt foreground illumination~\cite{guo2021intrinsic,careaga2023intrinsic}.
Diffusion-based approaches use environment lighting or rendered shading cues to control illumination with generative priors~\cite{jin2024neuralgaffer,kocsis2024lightit,zeng2024dilightnet}.
IC-Light~\cite{zhang2025iclight} imposes light-transport consistency to preserve intrinsic appearance under illumination changes.
It enforces consistency between combining lighting conditions and combining the corresponding object appearances during training.
Relightful Harmonization~\cite{ren2024relightful} connects relighting with harmonization by conditioning portrait illumination on the destination background.
Joint methods further couple foreground adaptation with shadow synthesis~\cite{zhou2024foreground,yu2024cfdiffusion,wang2026lgi}.
However, these methods still struggle to produce consistently realistic results under complex lighting and material conditions, limiting their generalization to diverse real-world scenes.

\subsection{Image Insertion and Composition}
Reference-based insertion uses generative models to adapt an object to a new scene~\cite{yang2023paintbyexample}.
Object-conditioned diffusion methods preserve identity through reference features while allowing changes in geometry and appearance~\cite{song2023objectstitch,zhang2023controlcom,chen2024anydoor,song2024imprint}.
Training-free alternatives reuse pretrained diffusion priors through image inversion or per-image optimization~\cite{lu2023tficon,chen2024freecompose}.
Insert Anything~\cite{song2026insert} formulates insertion as in-context editing in a diffusion transformer, supporting flexible object placement and appearance control.
For composition from a fixed paste, however, the user's chosen pose and visible details are input constraints rather than attributes to regenerate.

Composition also requires modeling how an object affects its surroundings.
Approaches based on counterfactual supervision, region-specific controls, or generative priors model shadows, reflections, and other scene effects~\cite{winter2024objectdrop,wang2025psdiffusion,lu2026shine}.
OmniPaint~\cite{yu2025omnipaint} treats insertion and removal as complementary tasks, using paired data and cycle-consistent refinement to learn both object appearance and environmental effects.
MV-CoLight~\cite{ren2025mvcolight} pursues efficient, consistent appearance across 2D and multiview compositions through feed-forward prediction rather than per-scene optimization.
Geometry-aware methods instead use reconstructed assets, rendering, or spatial proxies to control object--scene interactions~\cite{liang2024inverse,zhang2025zerocomp,gao2026comgs,gong2026direct}.
Our setting instead starts from a user-specified 2D paste without requiring 3D assets or geometric proxies.
To the best of our knowledge, image composition research dedicated to preserving fixed pasted content and fine details while modeling object--scene interactions remains scarce.
Through a clearer task definition and open resources, our work aims to support systematic research and evaluation in this setting.

\section{Dataset and Benchmark}
\label{sec:dataset}

\begin{figure*}[t]
    \centering
    \includegraphics[width=\textwidth]{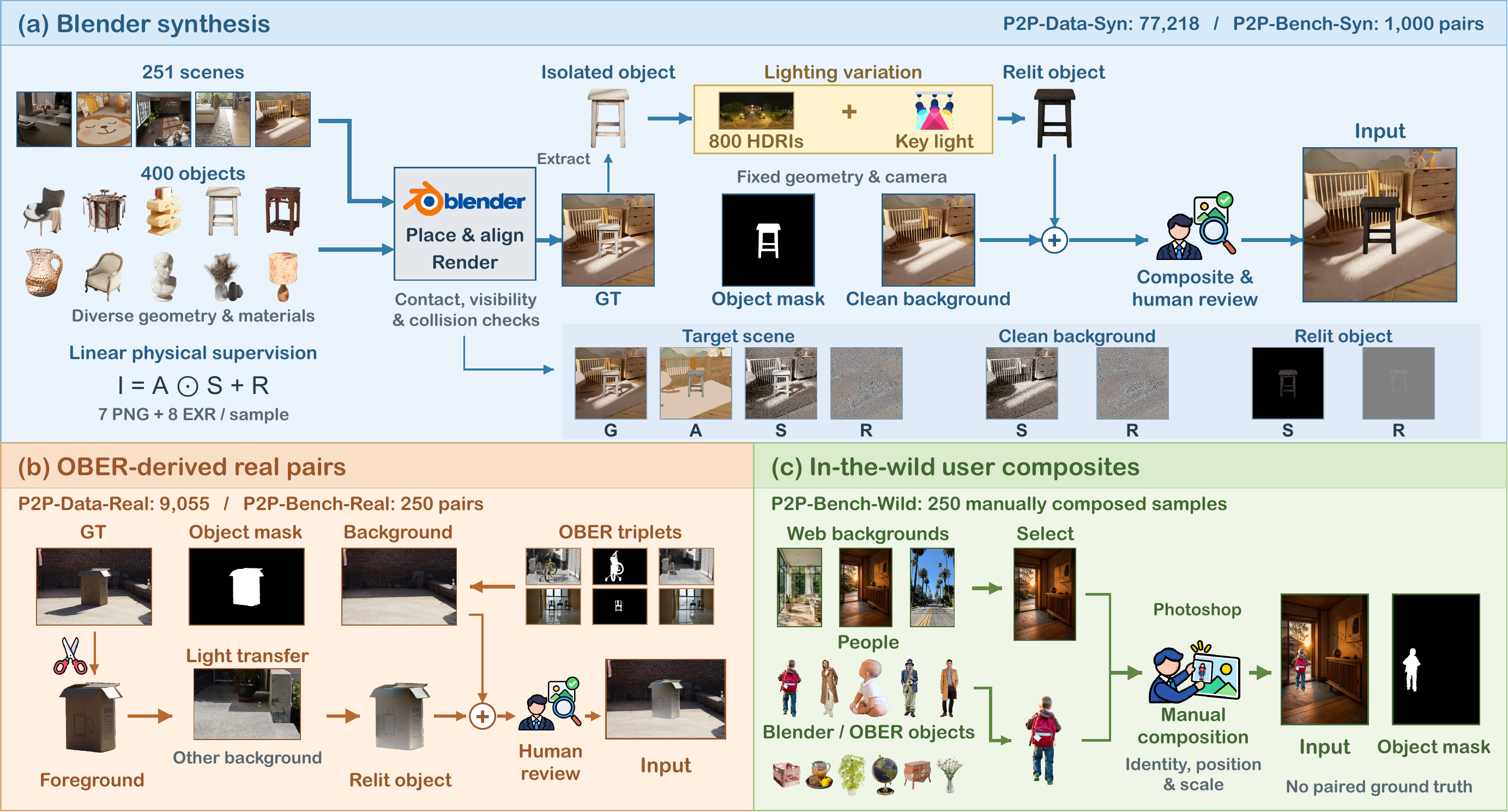}
    \caption{\textbf{Construction of P2P-Data and P2P-Bench.}
    (a) Controlled Blender rendering provides aligned composition pairs and physical annotations.
    (b) OBER-derived pairs match appearance-perturbed inputs to the original photographs.
    (c) Manually pasted foregrounds in new scenes form the unpaired Wild benchmark.}
    \label{fig:data-pipeline}
\end{figure*}

We construct P2P-Data with 86{,}273 training pairs and P2P-Bench with 1{,}500 evaluation examples (Fig.~\ref{fig:data-pipeline}).
Every example contains a pasted RGB input and an object mask.
Synthetic pairs provide aligned RGB targets and dense physical annotations, while OBER-derived pairs extend training to photographic content.
The benchmark combines held-out pairs with manually pasted inputs that simulate user editing.

\subsection{Physically Annotated Synthetic Pairs}
\label{sec:data-syn}

Each synthetic pair preserves object pose and texture while varying foreground illumination and object-induced scene effects.
Our Blender pipeline produces three aligned renders: the complete scene $G$, its object-free background $B$ under the same scene illumination, and the isolated object $O$ under different illumination (Fig.~\ref{fig:data-pipeline}(a)).
Camera, object pose, and background geometry are held fixed.
We illuminate $O$ with an HDR environment and a key light, varying their direction, intensity, and color.
Pasting $O$ onto $B$ with the rendered object mask gives the input, while $G$ supplies the target.

The aligned renders also provide the physical components underlying these appearance changes.
For $G$, we retain scene-linear RGB, diffuse albedo, diffuse shading, and residual appearance; $B$ and $O$ provide the corresponding shading and residual components.
Their differences supply supervision for object relighting and scene effects directly from rendering, rather than from intrinsic estimates of RGB images.
We retain the annotations in scene-linear HDR form for the physical targets defined in Sec.~\ref{sec:physical-targets}.

Our Blender pipeline combines 251 scenes, 400 object assets, and 800 HDR environment maps.
The assets include diffuse, glossy, metallic, transmissive, and emissive surfaces, producing diverse contact and cast shadows, reflections, and lighting effects.
Contact, visibility, and collision checks reject invalid placements, while render validation checks pair alignment and physical-component consistency.
Human review removes visibly implausible compositions, yielding 77{,}218 training pairs.

\subsection{Real-Scene Composition Pairs}
\label{sec:data-real}

P2P-Data-Real contains 9{,}055 pairs derived from the OBER object-removal dataset~\cite{zhao2026objectclear}.
Each OBER example provides an object-present image, its object-free background, and an object mask.
We turn removal into composition by retaining the object-present image as the target and pasting an appearance-perturbed foreground into its original object-free background (Fig.~\ref{fig:data-pipeline}(b)).

We perturb the foreground through Poisson blending~\cite{perez2003poisson} with another background, transferring color and brightness while preserving its spatial structure.
Only the perturbed foreground is returned to the original scene, at its original position and scale.
The resulting pair supervises foreground correction and restoration of scene effects against the original target.
After human review, these pairs provide photographic RGB supervision without physical component labels.

\begin{figure*}[!t]
    \centering
    \includegraphics[width=\textwidth]{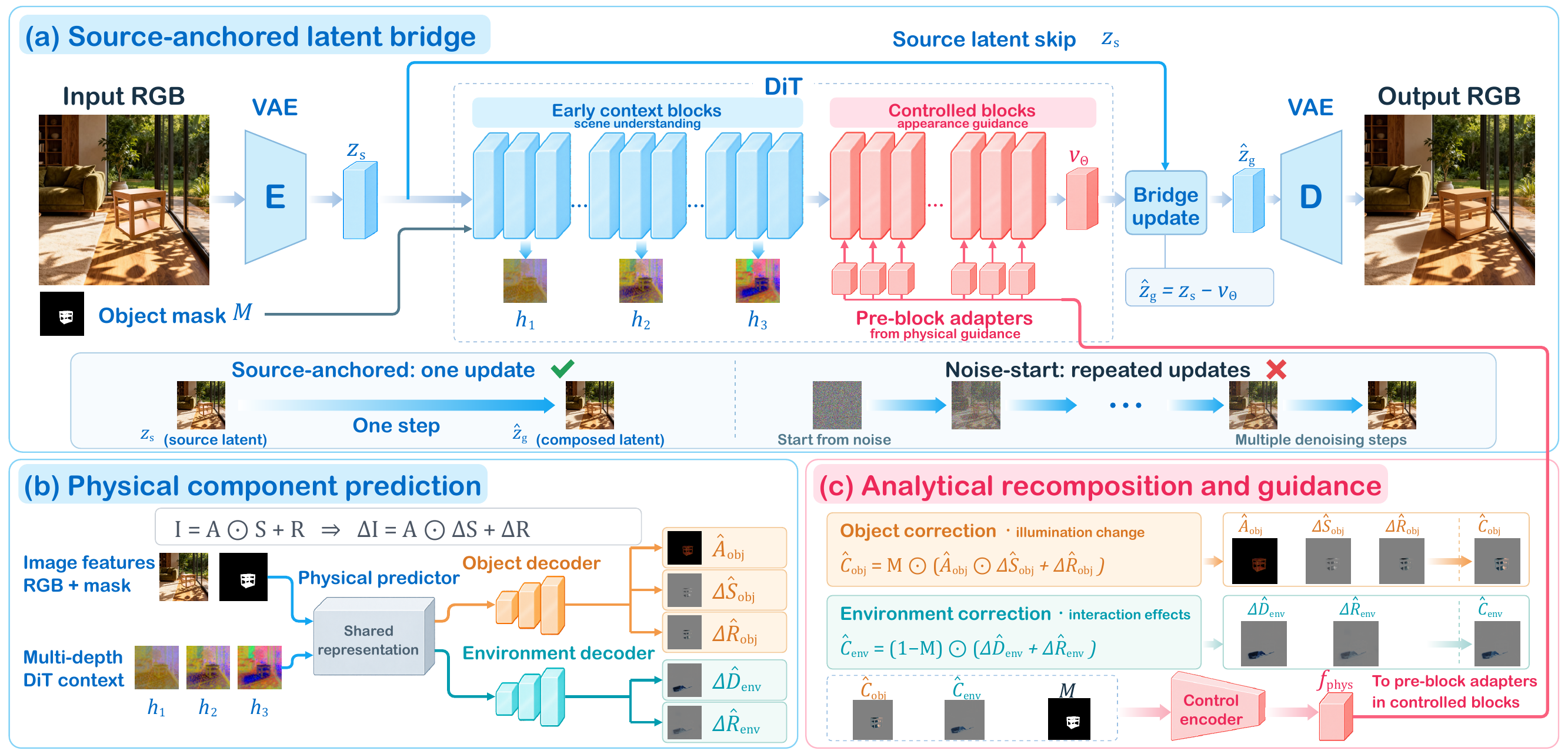}
    \caption{\textbf{Overview of Paste2Place.}
    (a) A source-anchored latent bridge maps the pasted input to the composite in one step.
    (b) Early DiT context and image--mask features support physical component prediction.
    (c) Analytic recomposition yields object and environment corrections that guide later DiT blocks within the same forward pass.}
    \label{fig:pipeline}
\end{figure*}

\subsection{Benchmark Construction}
\label{sec:data-bench}

P2P-Bench-Syn and P2P-Bench-Real contain 1{,}000 rendered and 250 OBER-derived pairs, respectively, manually selected to cover diverse object--scene arrangements, materials, and interaction types.
P2P-Bench-Syn uses scenes and HDR environment maps disjoint from those in P2P-Data-Syn.
Their aligned targets support full-reference evaluation of both foreground appearance and the effects of the pasted object on the surrounding scene.

P2P-Bench-Wild contains 250 manually pasted composites without paired ground truth.
We place person cutouts or objects from Blender and OBER into web-collected backgrounds, choosing their position and scale manually (Fig.~\ref{fig:data-pipeline}(c)).
Unlike the paired subsets, these inputs are created independently of a target image and reflect user-chosen arrangements in new scenes.
This subset supports qualitative evaluation and our user study.

\section{Methodology}
\label{sec:method}

Given a pasted RGB image $x_s$ and its object mask $M\in[0,1]$, \method maps the input to a realistic composite $\hat x_g$ in one end-to-end model.
The goal is to reconcile object illumination and scene effects while retaining the pasted arrangement and fine details.
We combine a source-anchored latent bridge with physical guidance for the required appearance changes (Fig.~\ref{fig:pipeline}).
P2P-Data provides aligned targets $x_g$ and synthetic physical annotations for learning this mapping.

\subsection{Source-Anchored Latent Bridge}
\label{sec:bridge}

Image composition changes how an object appears in a scene, not which object is present or where it is placed.
The pasted input already specifies the object's identity, pose, textures, and surrounding layout; what remains is to correct its illumination and generate the effects of its presence.
Even with the pasted image as a condition, noise-to-image generation must reconstruct its specified content from a noise-initialized state.
We instead make the pasted image itself the starting state, so generation learns to transform the existing content into a coherent composite (Fig.~\ref{fig:pipeline}(a)).
We formulate composition as source-anchored latent transport, inspired by image-to-image bridge models~\cite{chadebec2025lbm,tan2026vibt,liu2023i2sb,li2023bbdm}.

To exploit pretrained generative representations, we perform transport in VAE latent space and adapt a pretrained diffusion transformer to predict the update.
A frozen variational autoencoder with encoder $\mathcal E$ and decoder $\mathcal D$ maps the pasted image and its paired target to $z_s=\mathcal E(x_s)$ and $z_g=\mathcal E(x_g)$.
The aligned endpoints define a stochastic bridge $z_\tau=(1-\tau)z_s+\tau z_g+\sigma\sqrt{\tau(1-\tau)}\,\epsilon$, where $\epsilon\sim\mathcal N(0,I)$ and $\sigma$ controls the noise level.
The resized object mask is concatenated with the source representation to identify the inserted region.

During training, we sample $\tau\in[0,1)$ along the bridge and train the physically guided model $v_\Theta$ with the bridge-matching objective:
\begin{equation}
 \mathcal L_{\mathrm{bridge}}
 =\mathbb E_{\tau,\epsilon}\!\left[
 \left\|v_\Theta(z_\tau,\tau,x_s,M)
 -\frac{z_\tau-z_g}{1-\tau}\right\|_2^2
 \right].
 \label{eq:bridge-loss}
\end{equation}
At inference, we evaluate the model at the source endpoint $\tau=0$.
We subtract its prediction from the source latent and decode the result:
\begin{equation}
 \hat z_g=z_s-v_\Theta(z_s,0,x_s,M),
 \qquad \hat x_g=\mathcal D(\hat z_g).
 \label{eq:latent-estimate}
\end{equation}
This produces the final composite with one DiT evaluation, without random initialization or iterative sampling.
Object relighting and scene effects are synthesized together within this source-to-target mapping, using only the pasted image and object mask at inference.

\begin{figure}[!t]
    \centering
    \includegraphics[width=\columnwidth]{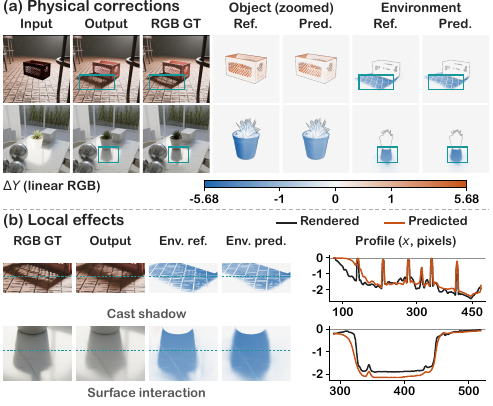}
    \caption{\textbf{Predicted physical corrections.}
    (a) Object and environment corrections versus rendered references.
    (b) Local crops and profiles along dashed lines.
    Maps and profiles share an $\operatorname{asinh}(\Delta Y/s)$ scale for linear-RGB luminance changes $\Delta Y$.
    Map colors indicate attenuation (blue) and increase (orange).}
    \label{fig:physical-visualization}
\end{figure}

\subsection{Physics-Based Appearance Change Modeling}
\label{sec:physical-targets}

The bridge objective supervises latent transport, but does not explicitly distinguish lighting changes from changes to material appearance.
We therefore guide the mapping with physical corrections that describe how illumination changes the object and how its presence affects the surrounding scene.
Both arise from light transport in the same arrangement.
We use a diffuse-plus-residual image formation model, as in intrinsic compositing and relighting~\cite{careaga2023intrinsic,liang2026pilight,careaga2024colorful,careaga2025physicalrelighting}:
\begin{equation}
 I=A\odot S+R,
 \label{eq:formation}
\end{equation}
Here $I$ is scene-linear radiance, and $A$, $S$, and $R$ are rendered proxies for diffuse albedo, RGB shading, and residual appearance, respectively.
The shading includes direct and indirect diffuse illumination, while $R=I-A\odot S$ accommodates specular reflection, transmission, emission, and departures from the diffuse model.
For a surface with fixed material, changing its illumination gives $\Delta I=A\odot\Delta S+\Delta R$.
This separates the intrinsic surface color to preserve from the illumination-dependent changes to predict.

Using the aligned target scene $G$, object-free background $B$, and differently illuminated object $O$ from Sec.~\ref{sec:data-syn}, we define the scene-linear source and target as
\begin{equation}
 I_s=M\odot I^O+(1-M)\odot I^B,
 \qquad I_g=I^G.
 \label{eq:source-construction}
\end{equation}

Inside the object, the source and target differ in illumination while sharing the same material.
We describe the object correction with its albedo $A_{\mathrm{obj}}=A^G$ and two changes, $\Delta S_{\mathrm{obj}}=S^G-S^O$ and $\Delta R_{\mathrm{obj}}=R^G-R^O$.
Outside the object, the receiving surfaces keep their materials, but the object's presence changes their illumination and residual appearance.
We represent the diffuse response directly as a radiance difference:
\begin{equation}
 \begin{aligned}
 \Delta D_{\mathrm{env}}&=A^G\odot(S^G-S^B),\\
 \Delta R_{\mathrm{env}}&=R^G-R^B.
 \end{aligned}
 \label{eq:physical-targets}
\end{equation}
Predicting $\Delta D_{\mathrm{env}}$ directly avoids requiring a full-scene albedo estimate.
Using $M$ as object coverage, we recompose these fields into object and environment corrections:
\begin{equation}
 \begin{aligned}
 C_{\mathrm{obj}}
 &=M\odot\left(A_{\mathrm{obj}}\odot\Delta S_{\mathrm{obj}}
                      +\Delta R_{\mathrm{obj}}\right),\\
 C_{\mathrm{env}}
 &=(1-M)\odot\left(\Delta D_{\mathrm{env}}
                      +\Delta R_{\mathrm{env}}\right).
 \end{aligned}
 \label{eq:physical-compose}
\end{equation}
These signed corrections describe both attenuation and added radiance.
Diffuse changes account for shadowing and indirect illumination, while residual changes capture effects such as reflections and emitted light.
Figure~\ref{fig:physical-visualization}(a) shows a positive object correction for the crate and attenuation for the plant pot, together with darkening of their receiving surfaces.
The crops and profiles in Fig.~\ref{fig:physical-visualization}(b) reveal how the environment corrections capture cast shadows and reflections.

\begin{figure*}[!t]
    \centering
    \includegraphics[width=\textwidth]{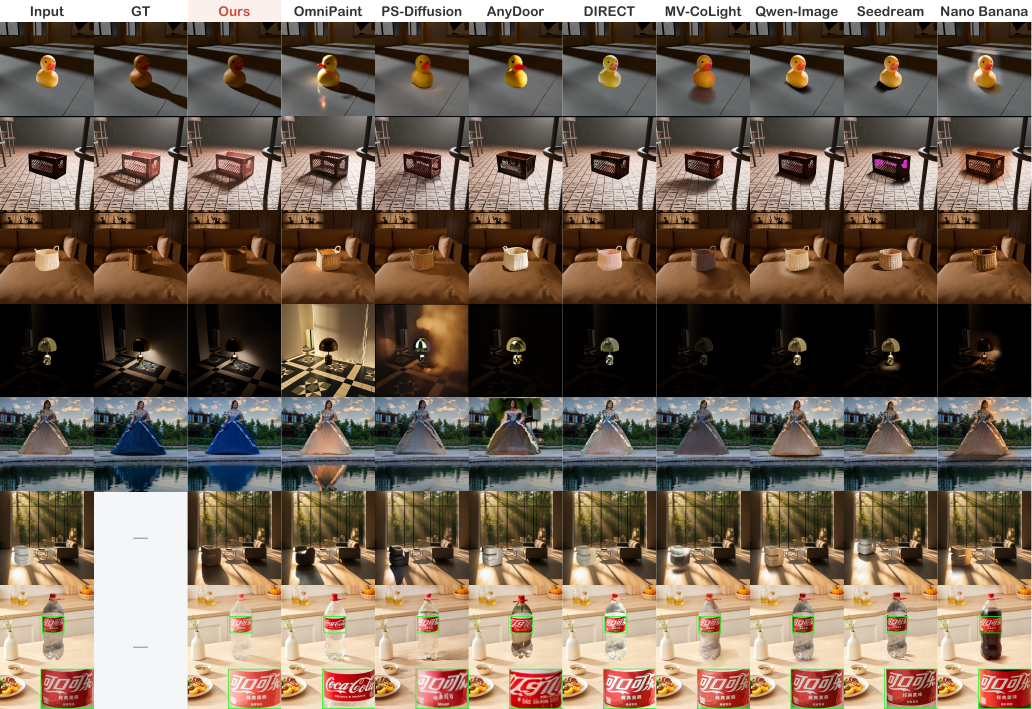}
    \caption{\textbf{Qualitative comparison on P2P-Bench.}
    The examples highlight differences in object relighting, scene effects, and fine-detail preservation.
    Green boxes mark the enlarged bottle-label regions.}
    \label{fig:qualitative-comparison}
\end{figure*}

\subsection{Physics-Guided Composition Framework}
\label{sec:framework}

We integrate physical prediction and control within the same DiT evaluation: early features estimate the required changes, and their recomposed corrections guide the later blocks.
Predicting these changes requires both scene context and local image structure.
The Physical Predictor therefore combines multi-depth DiT features with a pixel-aligned RGB--mask pyramid~\cite{ranftl2021dpt}, using shared representations and separate object and environment decoders (Fig.~\ref{fig:pipeline}(b)).

The Physical Composer converts the predicted fields into linear-RGB corrections through Eq.~\eqref{eq:physical-compose}.
After asinh compression of their HDR range, the Controller encodes both corrections $\hat C_{\mathrm{obj}}$ and $\hat C_{\mathrm{env}}$ together with $M$.
Layer-specific adapters inject these features into the image tokens before the later DiT blocks (Fig.~\ref{fig:pipeline}(c)).
Analytical recomposition specifies the required RGB changes, while the Controller learns how these changes should guide the latent update.

Synthetic physical supervision ties the guidance to the rendered appearance changes.
On synthetic pairs, $\mathcal L_{\mathrm{field}}$ supervises the five physical quantities separately, limiting compensation between albedo, shading, and residual errors.
$\mathcal L_{\mathrm{comp}}$ supervises their recomposed sum $\hat C_{\mathrm{obj}}+\hat C_{\mathrm{env}}$ against the rendered reference $C_{\mathrm{obj}}+C_{\mathrm{env}}$.
Signed fields and recomposed corrections are compared after asinh compression, using region-normalized Charbonnier penalties with higher weight on stronger environment changes.
$\mathcal L_{\mathrm{grad}}$ matches spatial gradients of the compressed $\Delta D_{\mathrm{env}}$ at multiple scales to supervise diffuse effect structure.
Together with the bridge objective, these terms give
\begin{equation}
 \mathcal L=\mathcal L_{\mathrm{bridge}}
 +\lambda_{\mathrm{phys}}\!
 \left(\mathcal L_{\mathrm{field}}
 +\lambda_c\mathcal L_{\mathrm{comp}}
 +\lambda_g\mathcal L_{\mathrm{grad}}\right),
 \label{eq:total-loss}
\end{equation}
Object terms are evaluated on $M$ and environment terms on $1-M$.
Real-scene pairs contribute the bridge loss without physical labels.
During joint training, the bridge loss also updates the Predictor and Controller, allowing paired RGB supervision to shape the physical guidance.

Training first establishes physical prediction and the source-to-target mapping with control disabled, using synthetic supervision before incorporating real-scene pairs.
We then train the Controller with ground-truth physical fields and adapt the Predictor through the trained Controller.
Finally, we jointly fine-tune the model with Eq.~\eqref{eq:total-loss}, using its own physical predictions for guidance.

\section{Experiments}
\label{sec:experiments}

\begin{table*}[!t]
    \centering
    \caption{\textbf{Quantitative comparison on P2P-Bench.}
    \textbf{Bold} and \underline{underlined} values indicate the best and second-best reported results.}
    \label{tab:main-comparison}
    \begingroup
\setlength{\tabcolsep}{1.6pt}
\renewcommand{\arraystretch}{1.25}
\fontsize{7}{8.2}\selectfont
\begin{tabular*}{\textwidth}{@{\extracolsep{\fill}}l*{12}{c}@{}}
\toprule
Metric & MV-CoLight & OmniPaint & PS-Diffusion & DreamFuse & AnyDoor & DIRECT & SHINE & Insert Anything & Qwen-Image & Seedream & Nano Banana & \textbf{Ours} \\
\midrule
PSNR $\uparrow$ & \underline{24.35} & 21.67 & 21.27 & 23.70 & 20.43 & 22.01 & 20.24 & 21.84 & 22.32 & 22.11 & 23.03 & \textbf{30.83} \\
SSIM $\uparrow$ & \underline{0.9376} & 0.8672 & 0.7830 & 0.9358 & 0.8262 & 0.9344 & 0.7900 & 0.9009 & 0.9261 & 0.9212 & 0.9290 & \textbf{0.9427} \\
LPIPS $\downarrow$ & 0.0686 & 0.1043 & 0.2108 & \underline{0.0665} & 0.1618 & 0.0687 & 0.1859 & 0.0976 & 0.0678 & 0.0725 & 0.0687 & \textbf{0.0396} \\
fMSE $\downarrow$ & \underline{3091.1} & 5744.9 & 3986.7 & 3640.7 & 6594.0 & 6700.9 & 6543.0 & 6026.6 & 5853.4 & 5773.6 & 4041.5 & \textbf{713.2} \\
Effect MAE $\downarrow$ & \underline{0.1343} & 0.1344 & 0.1414 & 0.1571 & 0.1676 & 0.1687 & 0.1709 & 0.1722 & 0.1433 & 0.1504 & 0.1374 & \textbf{0.0589} \\
\midrule
Latency (s) $\downarrow$ & \underline{1.289} & 20.415 & 16.795 & 41.518 & 8.500 & 70.257 & 115.424 & 22.661 & 15.677 & 65.619 & 27.559 & \textbf{0.743} \\
\bottomrule
\end{tabular*}
\endgroup

\end{table*}

\subsection{Experimental Setup}
\label{sec:experimental-setup}

\paragraph{Evaluation data.}
We evaluate on P2P-Bench (Sec.~\ref{sec:data-bench}).
Its 1,000 Syn and 250 Real pairs support full-reference quantitative evaluation and qualitative comparison.
The 250 unpaired Wild composites represent user-created inputs and support qualitative evaluation and our user study.

\paragraph{Compared methods.}
We compare with 11 representative methods for lighting-aware composition, reference-based insertion, and general image editing:
MV-CoLight~\cite{ren2025mvcolight}, OmniPaint~\cite{yu2025omnipaint}, PS-Diffusion~\cite{wang2025psdiffusion}, DreamFuse~\cite{huang2025dreamfuse}, AnyDoor~\cite{chen2024anydoor}, DIRECT~\cite{gong2026direct}, SHINE~\cite{lu2026shine}, Insert Anything~\cite{song2026insert}, Qwen-Image~2.1~\cite{qwen2026image21}, Seedream~5.0~Lite~\cite{bytedance2026seedream5}, \mbox{Nano Banana Pro}~\cite{google2025nanobananapro}.

\paragraph{Evaluation metrics.}
We use PSNR and SSIM~\cite{wang2004ssim} to assess reconstruction fidelity, and LPIPS~\cite{zhang2018lpips} to measure perceptual similarity.
Foreground MSE (fMSE) evaluates appearance within the object mask, while effect MAE measures error in background regions that change between the pasted input and the target.
These regional metrics focus on the object and its effects rather than the unchanged background.

\paragraph{Implementation.}
We use the FLUX.2 Klein 9B transformer~\cite{blackforestlabs2026flux2} with a frozen VAE.
Training on P2P-Data updates rank-256 LoRA adapters~\cite{hu2022lora}, selected backbone layers, and the Physical Predictor and Controller.
Syn and Real images are resized to $768\times768$ and $480\times640$ (height $\times$ width), respectively.
We train with AdamW for 215,000 updates across the training stages described in Sec.~\ref{sec:framework}, using a total batch size of four and BF16 on two NVIDIA H100 GPUs (80\,GB of memory each).
LoRA and physical-module learning rates start at $4\times10^{-5}$ and $10^{-4}$, respectively, and are halved for final joint refinement.
Our results are generated in one step at the input resolution without text prompts.

\subsection{Comparison with State-of-the-Art Methods}
\label{sec:comparison}

\paragraph{Quantitative results.}
Table~\ref{tab:main-comparison} shows that \method achieves the best scores on all five quality metrics.
Against the strongest baseline for each metric, it improves PSNR by 6.48\,dB over MV-CoLight and reduces LPIPS by 40.4\% relative to DreamFuse.
MV-CoLight and DreamFuse also attain SSIM above 0.93, but their foreground and effect-region errors remain markedly higher than ours.
Foreground MSE falls from 3,091.1 to 713.2 and effect MAE from 0.1343 to 0.0589, showing that the gains cover both the inserted object and its surrounding effects.
This distinction matters because full-image similarity can be dominated by background regions that require no change, whereas composition depends on correcting the object and the regions it affects.

\label{sec:efficiency}
Alongside these quality gains, the complete one-step model averages 0.743\,s end to end on a single H100 80\,GB GPU, the lowest reported latency in Table~\ref{tab:main-comparison}.
This includes physical prediction and control without iterative sampling, giving users faster feedback when comparing placements.
For Nano Banana Pro and Seedream, the reported times also include API service latency.

\paragraph{Qualitative results.}
Figure~\ref{fig:qualitative-comparison} compares object relighting, scene effects, and preservation of structure and fine details across diverse scenes.
In the first row, \method changes object shading together with its directional cast shadow.
Several alternatives retain a bright foreground or add only a small contact shadow.
The lamp instead requires additional illumination: our localized lighting pattern is closer to the reference than outputs that brighten the whole room or leave it largely unchanged.
These two cases require opposite changes in scene brightness, from darkening caused by cast shadows to added illumination from an emissive object.
These examples illustrate why a coherent composite requires changes on both sides of the object boundary.
The enlarged bottle-label regions in the last row compare preservation of fine text details.
\method preserves the lettering while adapting the bottle's appearance to the scene; several alternatives replace characters or blur their strokes.

\subsection{User Study}
\label{sec:user-study}

We conduct a user study with 15 volunteers on the 250 P2P-Bench-Wild composites, collecting their ratings across the 12 methods.
Participants assess overall composition quality on a 1--5 scale from poor (1) to excellent (5), considering subject fidelity, natural integration, and lighting consistency.

\begin{figure}[!ht]
    \centering
    \includegraphics[width=\columnwidth]{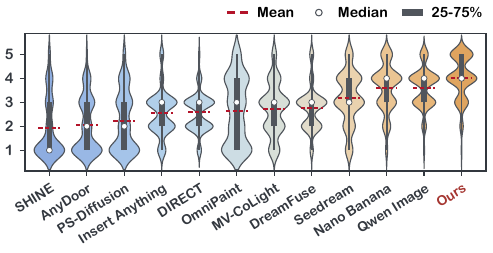}
    \caption{\textbf{Human ratings of composition quality.}
    Methods are ordered by mean score; violin widths show rating density.}
    \label{fig:user-study}
\end{figure}

Figure~\ref{fig:user-study} shows that \method obtains the highest mean score, 4.01 out of 5, compared with 3.60 for Qwen-Image and 3.58 for Nano Banana Pro.
For \method, 75.1\% of ratings are good or excellent, compared with 57.4\% for Qwen-Image and 58.6\% for Nano Banana Pro.
These ratings support the gains observed in the paired evaluation.

\subsection{Ablation Studies}
\label{sec:ablations}

Table~\ref{tab:ablation} examines task-specific bridge adaptation and the contribution of physical guidance on P2P-Bench.

\begin{table}[!ht]
    \centering
    \caption{\textbf{Bridge adaptation and physical guidance.}
    Results on 1,250 paired examples compare FLUX.2 9B fine-tuned on P2P-Data with \method variants.}
    \label{tab:ablation}
    \begingroup
\setlength{\tabcolsep}{1.6pt}
\renewcommand{\arraystretch}{1.1}
\fontsize{7.5}{9}\selectfont
\begin{tabular*}{\columnwidth}{@{\extracolsep{\fill}}lccccc@{}}
\toprule
Setting & PSNR $\uparrow$ & SSIM $\uparrow$ & LPIPS $\downarrow$ & fMSE $\downarrow$ & Effect MAE $\downarrow$ \\
\midrule
FLUX.2 9B (fine-tuned) & 24.33 & 0.8645 & 0.0742 & 1649.3 & 0.0980 \\
\midrule
w/o physical branch & 28.58 & 0.9265 & 0.0493 & 841.9 & 0.0725 \\
Spatially shifted guidance & 28.20 & 0.9298 & 0.0584 & 905.6 & 0.0801 \\
\textbf{Ours} & \textbf{30.83} & \textbf{0.9427} & \textbf{0.0396} & \textbf{713.2} & \textbf{0.0589} \\
\bottomrule
\end{tabular*}
\endgroup

\end{table}

\paragraph{Source-anchored latent bridge.}
With both models trained on P2P-Data, \method without the physical branch outperforms fine-tuned FLUX.2 9B on all five metrics.
PSNR increases from 24.33 to 28.58\,dB and LPIPS decreases from 0.0742 to 0.0493, indicating improvements in both reconstruction fidelity and perceptual similarity.
Lower foreground and effect-region errors show that the source-anchored model more accurately reproduces the object's appearance and its effects on the surrounding scene.
The full one-step model with physical guidance also reduces mean latency from 1.129\,s for fine-tuned FLUX.2 9B to 0.743\,s.

\paragraph{Physical guidance.}
The full model improves all five metrics over the variant trained independently without the physical branch: PSNR rises from 28.58 to 30.83\,dB and LPIPS falls from 0.0493 to 0.0396.
We also test the full model with correction maps circularly shifted by one third of their height and width, leaving its weights, input, and mask unchanged.
Shifting preserves the correction values but breaks their spatial correspondence with image content.
Compared with aligned guidance, fMSE increases from 713.2 to 905.6 and effect MAE from 0.0589 to 0.0801.
Together, these comparisons show that the physical branch improves composition through corrections that are spatially aligned with the scene.

\section{Conclusion}
\label{sec:conclusion}

We presented \method, an end-to-end model that turns pasted inputs into realistic composites in one step.
Its source-anchored latent bridge transforms existing content, guided by object and environment corrections analytically recomposed from predicted physical components.
This couples physical appearance modeling with preservation of the pasted layout and fine details.
P2P-Data provides synthetic and real-scene training pairs, with dense physical annotations for rendered examples; P2P-Bench covers paired and manually pasted composites.
Quantitative comparisons and human ratings show improved composition quality, while the complete model achieves sub-second mean end-to-end latency.
Our task definition and unified solution provide a foundation for further research on image composition.

{
    \small
    \bibliographystyle{ieeenat_fullname}
    \bibliography{main}
}
\end{document}